# Pixelwise Instance Segmentation with a Dynamically Instantiated Network

Anurag Arnab and Philip H.S Torr
University of Oxford
{anurag.arnab, philip.torr}@eng.ox.ac.uk

## Abstract

*Semantic segmentation and object detection research have recently achieved rapid progress. However, the former task has no notion of different instances of the same object, and the latter operates at a coarse, bounding-box level. We propose an Instance Segmentation system that produces a segmentation map where each pixel is assigned an object class and instance identity label. Most approaches adapt object detectors to produce segments instead of boxes. In contrast, our method is based on an initial semantic segmentation module, which feeds into an instance subnetwork. This subnetwork uses the initial category-level segmentation, along with cues from the output of an object detector, within an end-to-end CRF to predict instances. This part of our model is dynamically instantiated to produce a variable number of instances per image. Our end-to-end approach requires no post-processing and considers the image holistically, instead of processing independent proposals. Therefore, unlike some related work, a pixel cannot belong to multiple instances. Furthermore, far more precise segmentations are achieved, as shown by our substantial improvements at high $AP^r$ thresholds.*

## 1. Introduction

Semantic segmentation and object detection are well-studied scene understanding problems, and have recently witnessed great progress due to deep learning [22, 13, 7]. However, semantic segmentation – which labels every pixel in an image with its object class – has no notion of different instances of an object (Fig. 1). Object detection does localise different object instances, but does so at a very coarse, bounding-box level. Instance segmentation localises objects at a pixel level, as shown in Fig. 1, and can be thought of being at the intersection of these two scene understanding tasks. Unlike the former, it knows about different instances of the same object, and unlike the latter, it operates at a pixel level. Accurate recognition and localisation of objects enables many applications, such as autonomous driving [9], image-editing [53] and robotics [17].

Many recent approaches to instance segmentation are based on object detection pipelines where objects are first localised with bounding boxes. Thereafter, each bounding box is refined into a segmentation [19, 20, 32, 37, 30]. Another related approach [12, 56] is to use segment-based region proposals [10, 41, 42] instead of box-based proposals. However, these methods do not consider the entire image, but rather independent proposals. As a result, occlusions between different objects are not handled. Furthermore, many of these methods cannot easily produce segmentation maps of the image, as shown in Fig. 1, since they process numerous proposals independently. There are typically far more proposals than actual objects in the image, and these proposals can overlap and be assigned different class labels. Finally, as these methods are based on an initial detection step, they cannot recover from false detections.

Our proposed method is inspired by the fact that instance segmentation can be viewed as a more complex form of semantic segmentation, since we are not only required to label the object class of each pixel, but also its instance identity. We produce a pixelwise segmentation of the image, where each pixel is assigned both a semantic class and instance label. Our end-to-end trained network, which outputs a variable number of instances per input image, begins with an initial semantic segmentation module. The following, dynamic part of the network, then uses information from an object detector and a Conditional Random Field (CRF) model to distinguish different instances. This approach is robust to false-positive detections, as well as poorly localised bounding boxes which do not cover the entire object, in contrast to detection-based methods to instance segmentation. Moreover, as it considers the entire image when making predictions, it attempts to resolve occlusions between different objects and can produce segmentation maps as in Fig. 1 without any post-processing.

Furthermore, we note that the Average Precision (AP) metric [14] used in evaluating object detection systems, and its $AP^r$ variant [19] used for instance segmentation, considers individual, potentially overlapping, object predictions in isolation, as opposed to the entire image. To evaluate methods such as ours, which produce complete segmentation maps and reason about occlusions, we also evaluate using

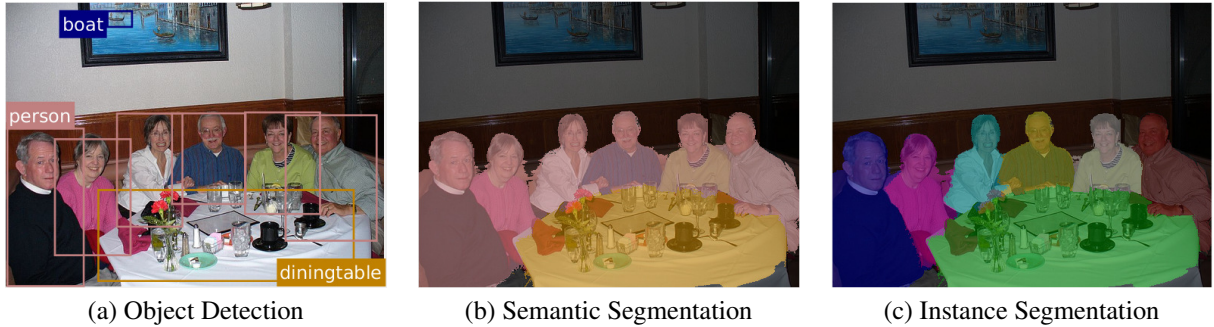

| (a) Object Detection | (b) Semantic Segmentation | (c) Instance Segmentation |

Figure 1: Object detection (a) localises the different people, but at a coarse, bounding-box level. Semantic segmentation (b) labels every pixel, but has no notion of instances. Instance segmentation (c) labels each pixel of each person uniquely. Our proposed method jointly produces both semantic and instance segmentations. Our method uses the output of an object detector as a cue to identify instances, but is robust to false positive detections, poor bounding box localisation and occlusions. Best viewed in colour.

the "Matching Intersection over Union" metric.

Our system, which is based on an initial semantic segmentation subnetwork, produces sharp and accurate instance segmentations. This is reflected by the substantial improvements we achieve over state-of-the-art methods at high $AP^r$ thresholds on the Pascal VOC and Semantic Boundaries datasets. Furthermore, our network improves on the semantic segmentation task while being trained for the related task of instance segmentation.

## 2. Related Work

An early work on instance segmentation was by Winn and Shotton [51]. A per-pixel unary classifier was trained to predict parts of an object. These parts were then encouraged to maintain a spatial ordering, that is characteristic of an instance, using asymmetric pairwise potentials in a Conditional Random Field (CRF). Subsequent work [54], presented another approach where detection outputs of DPM [15], with associated foreground masks, were assigned a depth ordering using a generative, probabilistic model. This depth ordering resolved occlusions.

However, instance segmentation has become more common after the "Simultaneous Detection and Segmentation" (SDS) work of Hariharan *et al*. [19]. This system was based on the R-CNN pipeline [16]: Region proposals, generated by the method of [1], were classified into object categories with a Convolutional Neural Network (CNN) before applying bounding-box regression as post-processing. A class-specific segmentation was then performed in this bounding box to simultaneously detect and segment the object. Numerous works [20, 8, 30] have extended this pipeline. However, approaches that segment instances by refining detections [19, 20, 8, 11, 30] are inherently limited by the quality of the initial proposals. This problem is exacerbated by the fact that this pipeline consists of several different modules trained with different objective functions. Furthermore, numerous post-processing steps such as "superpixel projection" and rescoring are performed. Dai *et al*. [12] addressed some of these issues by designing one end-to-end trained network that generates box-proposals, creates foreground masks from these proposals and then classifies these masks. This network can be seen as an extension of the end-to-end Faster-RCNN [44] detection framework, which generates box-proposals and classifies them. Additionally, Liu *et al*. [37] formulated an end-to-end version of the SDS network [19], whilst [32] iteratively refined object proposals.

On a separate track, algorithms have also been developed that do not require object detectors. Zhang *et al*. [57, 58] segmented car instances by predicting the depth ordering of each pixel in the image. Unlike the previous detection-based approaches, this method reasoned globally about all instances in the image simultaneously (rather than individual proposals) with an MRF-based formulation. However, inference of this graphical model was not performed end-to-end as shown to be possible in [60, 2, 5, 34]. Furthermore, although this method does not use object detections, it is trained with ground truth depth and assumes a maximum of nine cars in an image. Predicting all the instances in an image simultaneously (rather than classifying individual proposals) requires a model to be able to handle a variable number of output instances per image. As a result, [45] proposed a Recurrent Neural Network (RNN) for this task. However, this model was only for a single object category. Our proposed method not only outputs a variable number of instances, but can also handle multiple object classes.

Liang *et al*. [33] developed another proposal-free method based on the semantic segmentation network of [6]. The category-level segmentation, along with CNN features, was used to predict instance-level bounding boxes. The number of instances of each class was also predicted to enable a final spectral clustering step. However, this additional information predicted by Liang's network could have been obtained from an object detector. Arnab *et al*. [3] also started with an initial semantic segmentation network [2], and combined this with the outputs of an object detector using a CRF to reason about instances. This method was not trained end-to-end though, and could not really recover from errors in bounding-box localisation or occlusion.

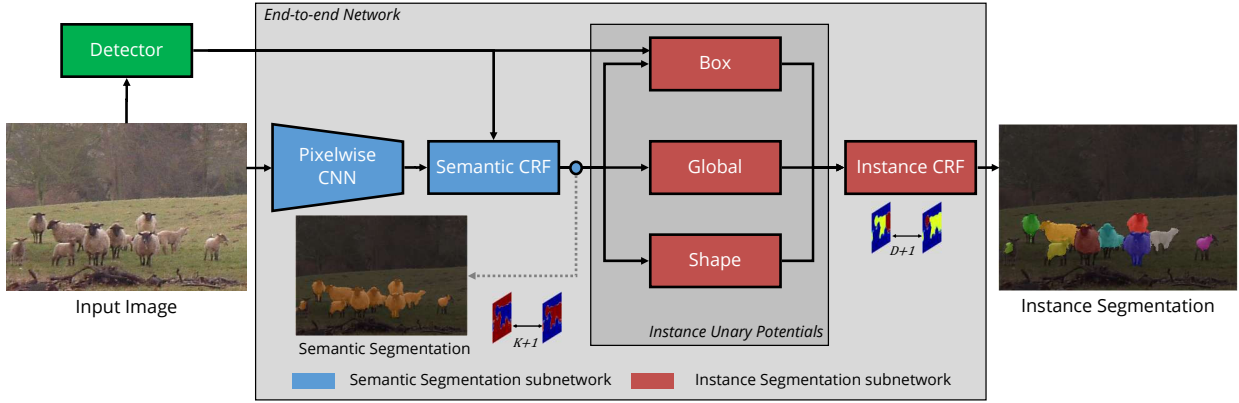

Figure 2: Network overview: Our end-to-end trained network consists of semantic- and instance-segmentation modules. The intermediate category-level segmentation, along with the outputs of an object detector, are used to reason about instances. This is done by instance unary terms which use information from the detector's bounding boxes, the initial semantic segmentation and also the object's shape. A final CRF is used to combine all this information together to obtain an instance segmentation. The output of the semantic segmentation module is a fixed size $W \times H \times (K + 1)$ tensor where $K$ is the number of object classes, excluding background, in the dataset. The final output, however, is of a variable $W \times H \times (D+1)$ dimensions where $D$ is the number of detected objects (and one background label).

Our method also has an initial semantic segmentation subnetwork, and uses the outputs of an object detector. However, in contrast to [3] it is trained end-to-end to improve on both semantic- and instance-segmentation performance (to our knowledge, this is the first work to achieve this). Furthermore, it can handle detector localisation errors and occlusions better due to the energy terms in our end-to-end CRF. In contrast to detection-based approaches [19, 20, 12, 37], our network requires no additional post-processing to create an instance segmentation map as in Fig. 1(c) and reasons about the entire image, rather than independent proposals. This global reasoning allows our method to produce more accurate segmentations. Our proposed system also handles a variable number of instances per image, and thus does not assume a maximum number of instances like [57, 58].

## 3. Proposed Approach

Our network (Fig. 2) contains an initial semantic segmentation module. We use the semantic segmentation result, along with the outputs of an object detector, to compute the unary potentials of a Conditional Random Field (CRF) defined over object instances. We perform mean field inference in this random field to obtain the Maximum a Posteriori (MAP) estimate, which is our labelling. Although our network consists of two conceptually different parts – a semantic segmentation module, and an instance segmentation network – the entire pipeline is fully differentiable, given object detections, and trained end-to-end.

### 3.1. Semantic Segmentation subnetwork

Semantic Segmentation assigns each pixel in an image a semantic class label from a given set, $\mathcal{L}$. In our case, this module uses the FCN8s architecture [38] which is based on the VGG [47] ImageNet model. For better segmentation results, we include mean field inference of a Conditional Random Field as the last layer of this module. This CRF contains the densely-connected pairwise potentials described in [26] and is formulated as a recurrent neural network as in [60]. Additionally, we include the Higher Order detection potential described in [2]. This detection potential has two primary benefits: Firstly, it improves semantic segmentation quality by encouraging consistency between object detections and segmentations. Secondly, it also recalibrates detection scores. This detection potential is similar to the one previously proposed by [28], [48], [52] and [55], but formulated for the differentiable mean field inference algorithm. We employ this potential as we are already using object detection information for identifying object instances in the next stage. We denote the output at the semantic segmentation module of our network as the tensor $\mathbf{Q}$, where $Q_i(l)$ denotes the probability (obtained by applying the softmax function on the network's activations) of pixel $i$ taking on the label $l \in \mathcal{L}$.

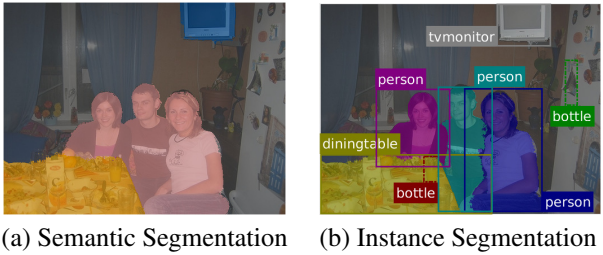

(a) Semantic Segmentation  (b) Instance Segmentation

Figure 3: Instance segmentation using only the "Box" unary potential. This potential is effective when we have a good initial semantic segmentation (a). Occlusions between objects of the same class can be resolved by the pairwise term based on appearance differences. Note that we can ignore the confident, false-positive "bottle" detections (b). This is in contrast to methods such as [8, 19, 20, 30] which cannot recover from detection errors.

### 3.2. Instance Segmentation subnetwork

At the input to our instance segmentation subnetwork, we assume that we have two inputs available: The semantic segmentation predictions, $\mathbf{Q}$, for each pixel and label, and a set of object detections. For each input image, we assume that there are $D$ object detections, and that the $i^{\text{th}}$ detection is of the form $(l_i, s_i, B_i)$ where $l_i \in \mathcal{L}$ is the detected class label, $s_i \in [0, 1]$ is the confidence score and $B_i$ is the set of indices of the pixels falling within the detector's bounding box. Note that the number $D$ varies for every input image.

The problem of instance segmentation can then be thought of as assigning every pixel to either a particular object detection, or the background label. This is based on the assumption that every object detection specifies a potential object instance. We define a multinomial random variable, $V$, at each of the $N$ pixels in the image, and $\mathbf{V} = [V_1 \, V_2 \ldots V_N]^T$. Each variable at pixel $i$, $V_i$, is assigned a label corresponding to its instance. This label set, $\{0, 1, 2, ..., D\}$ changes for each image since $D$, the number of detections, varies for every image (0 is the background label). In the case of instance segmentation of images, the quality of a prediction is invariant to the permutations of the instance labelling. For example, labelling the "blue person" in Fig. 1(c) as "1" and the "purple person" as "2" is no different to labelling them as "2" and "1" respectively. This condition is handled by our loss function in Sec. 3.4.

Note that unlike works such as [57] and [58] we do not assume a maximum number of possible instances and keep a fixed label set. Furthermore, since we are considering object detection outputs jointly with semantic segmentation predictions, we have some robustness to high-scoring false positive detections unlike methods such as [8, 20, 37] which refine object detections into segmentations.

We formulate a Conditional Random Field over our instance variables, $V$, which consists of unary and pairwise

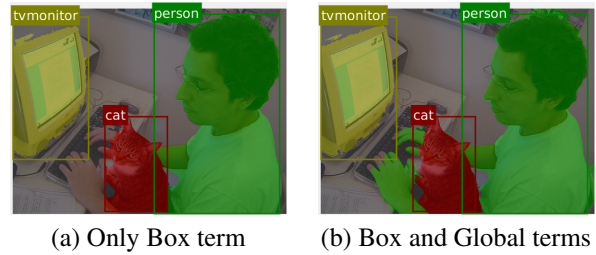

(a) Only Box term  (b) Box and Global terms

Figure 4: The "Global" unary potential (b) is particularly effective in cases where the input detection bounding box does not cover the entire extent of the object. Methods which are based on refining bounding-box detections such as [19, 20, 8, 12] cannot cope with poorly localised detections. Note, the overlaid detection boxes are an additional input to our system.

energies. The energy of the assignment $\mathbf{v}$ to all the variables, $\mathbf{V}$, is

$$E(\mathbf{V} = \mathbf{v}) = \sum_i U(v_i) + \sum_{i<j} P(v_i, v_j). \quad (1)$$

The unary energy is a sum of three terms, which take into account the object detection bounding boxes, the initial semantic segmentation and shape information,

$$U(v_i) = -\ln[w_1 \psi_{Box}(v_i) + w_2 \psi_{Global}(v_i) + w_3 \psi_{Shape}(v_i)], \quad (2)$$

and are described further in Sections 3.2.1 through 3.2.3. $w_1$, $w_2$ and $w_3$ are all weighting co-efficients learned via backpropagation.

#### 3.2.1 Box Term

This potential encourages a pixel to be assigned to the instance corresponding to the $k^{\text{th}}$ detection if it falls within the detection's bounding box. This potential is proportional to the probability of the pixel's semantic class being equal to the detected class $Q_i(l_k)$ and the detection score, $s_k$.

$$\psi_{Box}(V_i = k) = \begin{cases} Q_i(l_k) s_k & \text{if } i \in B_k \\ 0 & \text{otherwise} \end{cases} \quad (3)$$

As shown in Fig. 3, this potential performs well when the initial semantic segmentation is good. It is robust to false positive detections, unlike methods which refine bounding boxes [8, 19, 20] since the detections are considered in light of our initial semantic segmentation, $\mathbf{Q}$. Together with the pairwise term (Sec. 3.2.4), occlusions between objects of the same class can be resolved if there are appearance differences in the different instances.

### 3.2.2 Global Term

This term does not rely on bounding boxes, but only the segmentation prediction at a particular pixel, $Q_i$. It encodes the intuition that if we only know there are $d$ possible instances of a particular object class, and have no further localisation information, each instance is equally probable, and this potential is proportional to the semantic segmentation confidence for the detected object class at that pixel:

$$\psi_{Global}(V_i = k) = Q_i(l_k). \qquad (4)$$

As shown in Fig. 4, this potential overcomes cases where the bounding box does not cover the entire extent of the object, as it assigns probability mass to a particular instance label throughout all pixels in the image. This is also beneficial during training, as it ensures that the final output is dependent on the segmentation prediction at all pixels in the image, leading to error gradients that are more stable across batches and thus more amenable to backpropagation.

### 3.2.3 Shape Term

We also incorporate shape priors to help us reason about occlusions involving multiple objects of the same class, which may have minimal appearance variation between them, as shown in Fig. 5. In such cases, a prior on the expected shape of an object category can help us to identify the foreground instance within a bounding box. Previous approaches to incorporating shape priors in segmentation [23, 8, 50] have involved generating "shape exemplars" from the training dataset and, at inference time, matching these exemplars to object proposals using the Chamfer distance [46, 36].

We propose a fully differentiable method: Given a set of shape templates, $\mathcal{T}$, we warp each shape template using bilinear interpolation into $\tilde{\mathcal{T}}$ so that it matches the dimensions of the $k^{\text{th}}$ bounding box, $B_k$. We then select the shape prior which matches the segmentation prediction for the detected class within the bounding box, $\mathbf{Q}_{B_k}(l_k)$, the best according to the normalised cross correlation. Our shape prior is then the Hadamard (elementwise) product ($\odot$) between the segmentation unaries and the matched shape prior:

$$t^* = \arg\max_{t \in \tilde{\mathcal{T}}} \frac{\sum \mathbf{Q}_{B_k}(l_k) \odot t}{\|\mathbf{Q}_{B_k}(l_k)\| \, \|t\|} \qquad (5)$$

$$\psi(\mathbf{V}_{B_k} = k) = \mathbf{Q}_{B_k}(l_k) \odot t^*. \qquad (6)$$

Equations 5 and 6 can be seen as a special case of max-pooling, and the numerator of Eq. 5 is simply a convolution that produces a scalar output since the two arguments are of equal dimension. Additionally, during training, we can consider the shape priors $\mathcal{T}$ as parameters of our "shape term" layer and backpropagate through to the matched exemplar $t^*$ to update it. In practice, we initialised these parameters with the shape priors described in [50]. This consists of roughly 250 shape templates for each of five different aspect ratios. These were obtained by clustering foreground masks of object instances from the training set.

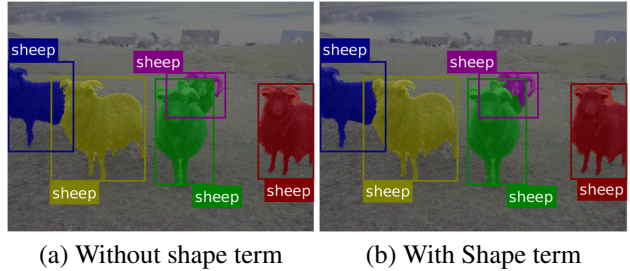

(a) Without shape term  (b) With Shape term

Figure 5: The "Shape" unary potential (b) helps us to distinguish between the green and purple sheep, which the other two unary potentials cannot. Input detections are overlaid on the images.

Here, we have only matched a single shape template to a proposed instance. This method could be extended in future to matching multiple templates to an instance, in which case each shape exemplar would correspond to a part of the object such as in DPM [15].

### 3.2.4 Pairwise term

The pairwise term consists of densely-connected Gaussian potentials [26] and encourages appearance and spatial consistency. The weights governing the importance of these terms are also learnt via backpropagation, as in [60]. We find that these priors are useful in the case of instance segmentation as well, since nearby pixels that have similar appearance often belong to the same object instance. They are often able to resolve occlusions based on appearance differences between objects of the same class (Fig. 3).

### 3.3. Inference of our Dynamic Instance CRF

We use mean field inference to approximately minimise the Gibbs Energy in Eq. 1 which corresponds to finding the Maximum a Posteriori (MAP) labelling of the corresponding probability distribution, $P(\mathbf{V} = \mathbf{v}) = \frac{1}{Z} \exp(-E(\mathbf{v}))$ where $Z$ is the normalisation factor. Mean field inference is differentiable, and this iterative algorithm can be unrolled and seen as a recurrent neural network [60]. Following this approach, we can incorporate mean field inference of a CRF as a layer of our neural network. This enables us to train our entire instance segmentation network end-to-end.

Because we deal with a variable number of instances for every image, our CRF needs to be dynamically instantiated to have a different number of labels for every image, as observed in [3]. Therefore, unlike [60], none of our weights are class-specific. This weight-sharing not only allows us to deal with variable length inputs, but class-specific weights also do not make sense in the case of instance segmentation since a class label has no particular semantic meaning.

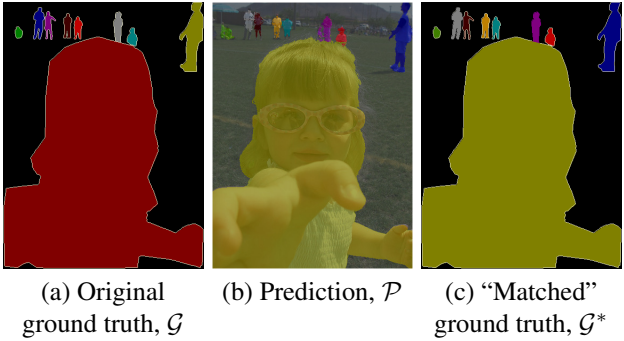

| (a) Original ground truth, $\mathcal{G}$ | (b) Prediction, $\mathcal{P}$ | (c) "Matched" ground truth, $\mathcal{G}^*$ |

Figure 6: Due to the problem of label permutations, we "match" the ground truth with our prediction before computing the loss when training.

### 3.4. Loss Function

When training for instance segmentation, we have a single loss function which we backpropagate through our instance- and semantic-segmentation modules to update all the parameters. As discussed previously, we need to deal with different permutations of our final labelling which could have the same final result. The works of [57] and [58] order instances by depth to break this symmetry. However, this requires ground-truth depth maps during training which we do not assume that we have. Proposal-based methods [12, 19, 20, 37] do not have this issue since they consider a single proposal at a time, rather than the entire image. Our approach is similar to [45] in that we match the original ground truth to our instance segmentation prediction based on the Intersection over Union (IoU) [14] of each instance prediction and ground truth, as shown in Fig. 6.

More formally, we denote the ground-truth labelling of an image, $\mathcal{G}$, to be a set of $r$ segments, $\{g_1, g_2, \ldots, g_r\}$, where each segment (set of pixels) is an object instance and has an associated semantic class label. Our prediction, which is the output of our network, $\mathcal{P}$, is a set of $s$ segments, $\{p_1, p_2, \ldots, p_s\}$, also where each segment corresponds to an instance label and also has an associated class label. Note that $r$ and $s$ may be different since we may predict greater or fewer instances than actually present. Let $\mathcal{M}$ denote the set of all permutations of the ground-truth, $\mathcal{G}$. As can be seen in Fig. 6, different permutations of the ground-truth correspond to the same qualitative result. We define the "matched" ground-truth, $\mathcal{G}^*$, as the permutation of the original ground-truth labelling which maximises the IoU between the prediction, $\mathcal{P}$, and ground truth:

$$\mathcal{G}^* = \arg\max_{m \in \mathcal{M}} \text{IoU}(m, \mathcal{P}). \quad (7)$$

Once we have the "matched" ground truth, $\mathcal{G}^*$, (Fig. 6) for an image, we can apply any loss function to train our network for segmentation. In our case, we use the common cross-entropy loss function. We found that this performed better than the approximate IoU loss proposed in [27, 45].

Crucially, we do not need to evaluate all permutations of the ground truth to compute Eq. 7, since it can be formulated as a maximum-weight bipartite matching problem. The edges in our bipartite graph connect ground-truth and predicted segments. The edge weights are given by the IoU between the ground truth and predicted segments if they share the same semantic class label, and zero otherwise. Leftover segments are matched to "dummy" nodes with zero overlap.

Additionally, the ordering of the instances in our network are actually determined by the object detector, which remains static during training. As a result, the ordering of our predictions does not fluctuate much during training – it only changes in cases where there are multiple detections overlapping an object.

### 3.5. Network Training

We first train a network for semantic segmentation with the standard cross-entropy loss. In our case, this network is FCN8s [38] with a CRF whose inference is unrolled as an RNN and trained end-to-end, as described in [60] and [2]. To this pretrained network, we append our instance segmentation subnetwork, and finetune with instance segmentation annotations and only the loss detailed in Sec. 3.4. For the semantic segmentation subnetwork, we train with an initial learning rate of $10^{-8}$, momentum of 0.9 and batch size of 20. The learning rate is low since we do not normalise the loss by the number of pixels. This is so that images with more pixels contribute a higher loss. The normalised learning rate is approximately $2 \times 10^{-3}$. When training our instance segmentation network as well, we lower the learning rate to $10^{-12}$ and use a batch size of 1 instead. Decreasing the batch size gave empirically better results. We also clipped gradients (a technique common in training RNNs [40]) with $\ell_2$ norms above $10^9$. This threshold was set by observing "normal" gradient magnitudes during training. The relatively high magnitude is due to the fact that our loss is not normalised. In our complete network, we have two CRF inference modules which are RNNs (one each in the semantic- and instance-segmentation subnetworks), and gradient clipping facilitated successful training.

### 3.6. Discussion

Our network is able to compute a semantic and instance segmentation of the input image in a single forward pass. We do not require any post-processing, such as the patch aggregation of [37], "mask-voting" of [12], "superpixel projection" of [19, 20, 30] or spectral clustering of [33]. The fact that we compute an initial semantic segmentation means that we have some robustness to errors in the object detector (Fig. 3). Furthermore, we are not necessarily limited by poorly localised object detections either (Fig. 4). Our CRF model allows us to reason about the entire image at a time, rather than consider independent object proposals, as done in [19, 20, 12, 37, 30]. Although we do not train our object detector jointly with the network, it also means that our segmentation network and object detector do not succumb to the same failure cases. Moreover, it ensures that our instance labelling does not "switch" often during training, which makes learning more stable. Finally, note that although we perform mean field inference of a CRF within our network, we do not optimise the CRF's likelihood, but rather a cross-entropy loss (Sec. 3.4).

## 4. Experimental Evaluation

Sections 4.1 to 4.6 describe our evaluation on the Pascal VOC Validation Set [14] and the Semantic Boundaries Dataset (SBD) [18] (which provides per-pixel annotations to 11355 previously unlaballed images from Pascal VOC). Section 4.7 details results on Cityscapes [9].

### 4.1. Experimental Details

We first train a network for semantic segmentation, therafter we finetune it to the task of instance segmentation, as described in Sec. 3.5. Our training data for the semantic segmentation pretraining consists of images from Pascal VOC [14], SBD [18] and Microsoft COCO [35]. Finally, when finetuning for instance segmentation, we use only training data from either the VOC dataset, or from the SBD dataset. We train separate models for evaluating on the VOC Validation Set, and the SBD Validation Set. In each case, we remove validation set images from the initial semantic segmentation pretraining set. We use the publicly available R-FCN object detection framework [13], and ensure that the images used to train the detector do not fall into our test sets for instance segmentation.

### 4.2. Evaluation Metrics

We report the mean Average Precision over regions ($AP^r$) as defined by [19]. The difference between $AP^r$ and the AP metric used in object detection [14] is that the Intersection over Union (IoU) is computed over predicted and ground-truth regions instead of bounding boxes. Furthermore, the standard AP metric uses an IoU threshold of 0.5 to determine whether a prediction is correct or not. Here, we use a variety of IoU thresholds since larger thresholds require more precise segmentations. Additionally, we report the $AP^r_{vol}$ which is the average of the $AP^r$ for 9 IoU thresholds ranging from 0.1 to 0.9 in increments of 0.1.

However, we also observe that the $AP^r$ metric requires an algorithm to produce a ranked list of segments and their object class. It does not require, nor evaluate, the ability of an algorithm to produce a globally coherent segmentation map of the image, for example Fig. 1c. To measure this, we propose the "Matching IoU" which matches the predicted image and ground truth, and then calculates the corresponding IoU as defined in [14]. This matching procedure is the same as described in Sec. 3.4. This measure was originally proposed in [54], but has not been used since in evaluating instance segmentation systems.

Table 1: The effect of the different CRF unary potentials, and end-to-end training with them, on the VOC 2012 Validation Set.

| | $AP^r$ | | | $AP^r_{vol}$ | match IoU |
|---|---|---|---|---|---|
| | 0.5 | 0.7 | 0.9 | | |
| Box Term (piecewise) | 60.0 | 47.3 | 21.2 | 54.9 | 42.6 |
| Box+Global (piecewise) | 59.1 | 46.1 | 23.4 | 54.6 | 43.0 |
| Box+Global+Shape (piecewise) | 59.5 | 46.4 | 23.3 | 55.2 | 44.8 |
| Box Term (end-to-end) | 60.7 | 47.4 | 24.6 | 56.2 | 46.9 |
| Box+Global (end-to-end) | 60.9 | 48.1 | **25.5** | 56.7 | 47.1 |
| Box+Global+Shape (end-to-end) | **61.7** | **48.6** | 25.1 | **57.5** | **48.3** |

### 4.3. Effect of Instance Potentials and End-to-End training

We first perform ablation studies on the VOC 2012 Validation set. This dataset, consisting of 1464 training and 1449 validation images has very high-quality annotations with detailed object delineations which makes it the most suited for evaluating pixel-level segmentations.

In Tab. 1, we examine the effect of each of our unary potentials in our Instance subnetwork on overall performance. Furthermore, we examine the effect of end-to-end training the entire network as opposed to piecewise training. Piecewise training refers to freezing the pretrained semantic segmentation subnetwork's weights and only optimising the instance segmentation subnetwork's parameters. Note that when training with only the "Box" (Eq. 3) unary potential and pairwise term, we also have to add in an additional "Background" detection which encompasses the entire image. Otherwise, we cannot classify the background label.

We can see that each unary potential improves overall instance segmentation results, both in terms of $AP^r_{vol}$ and the Matching IoU. The "Global" term (Eq. 4) shows particular improvement over the "Box" term at the high $AP^r$ threshold of 0.9. This is because it can overcome errors in bounding box localisation (Fig. 4) and leverage our semantic segmentation network's accurate predictions to produce precise labellings. The "Shape" term's improvement in the $AP^r_{vol}$ is primarily due to an improvement in the $AP^r$ at low thresholds. By using shape priors, we are able to recover instances which were occluded and missed out. End-to-end training also improves results at all $AP^r$ thresholds. Training with just the "Box" term shows a modest improvement in the $AP^r_{vol}$ of 1.3%. Training with the "Global" and "Shape" terms shows larger improvements of 2.1% and 2.3% respectively. This may be because the "Box" term only considers the semantic segmentation at parts of the image covered by object detections. Once we include the "Global" term, we consider the semantic segmentation over the entire image for the detected class. Training makes more efficient use of images, and error gradients are more stable in this case.

Table 2: Comparison of Instance Segmentation performance to recent methods on the VOC 2012 Validation Set

| Method | $AP^r$ | | | | | $AP^r_{vol}$ |
|---|---|---|---|---|---|---|
| | 0.5 | 0.6 | 0.7 | 0.8 | 0.9 | |
| SDS [19] | 43.8 | 34.5 | 21.3 | 8.7 | 0.9 | – |
| Chen *et al*. [8] | 46.3 | 38.2 | 27.0 | 13.5 | 2.6 | – |
| PFN [33] | 58.7 | 51.3 | 42.5 | 31.2 | 15.7 | 52.3 |
| Arnab *et al*. [3] | 58.3 | 52.4 | 45.4 | 34.9 | 20.1 | 53.1 |
| MPA 1-scale [37] | 60.3 | 54.6 | 45.9 | 34.3 | 17.3 | 54.5 |
| MPA 3-scale [37] | **62.1** | **56.6** | 47.4 | 36.1 | 18.5 | 56.5 |
| Ours | 61.7 | 55.5 | **48.6** | **39.5** | **25.1** | **57.5** |

Table 3: Comparison of Instance Segmentation performance on the SBD Dataset

| Method | $AP^r$ | | $AP^r_{vol}$ | match IoU |
|---|---|---|---|---|
| | 0.5 | 0.7 | | |
| SDS [19] | 49.7 | 25.3 | 41.4 | – |
| MPA 1-scale [37] | 55.5 | – | 48.3 | – |
| Hypercolumn [20] | 56.5 | 37.0 | – | – |
| IIS [30] | 60.1 | 38.7 | – | – |
| CFM [11] | 60.7 | 39.6 | – | – |
| Hypercolumn rescore [20] | 60.0 | 40.4 | – | – |
| MPA 3-scale rescore [37] | 61.8 | – | 52.0 | – |
| MNC [12] | 63.5 | 41.5 | – | 39.0 |
| MNC, Instance FCN [10] | 61.5 | 43.0 | – | – |
| IIS sp. projection, rescore [30] | **63.6** | 43.3 | – | – |
| Ours (piecewise) | 59.1 | 42.1 | 52.3 | 41.8 |
| Ours (end-to-end) | 62.0 | **44.8** | **55.4** | **47.3** |

### 4.4. Results on VOC Validation Set

We then compare our best instance segmentation model to recent methods on the VOC Validation Set in Tab. 2. The fact that our algorithm achieves the highest $AP^r$ at thresholds above 0.7 indicates that our method produces more detailed and accurate segmentations.

At an IoU threshold of 0.9, our improvement over the previous state-of-the-art (MPA [37]) is 6.6%, which is a relative improvement of 36%. Unlike [37, 19, 8], our network performs an initial semantic segmentation which may explain our more accurate segmentations. Other segmentation-based approaches, [3, 33] are not fully end-to-end trained. We also achieve the best $AP^r_{vol}$ of 57.5%. The relatively small difference in $AP^r_{vol}$ to MPA [37] despite large improvements at high IoU thresholds indicates that MPA performs better at low IoU thresholds. Proposal-based methods, such as [37, 19] are more likely to perform better at low IoU thresholds since they output more proposals than actual instances in an image (SDS evaluates 2000 proposals per image). Furthermore, note that whilst MPA takes 8.7s to process an image [37], our method requires approximately 1.5s on the same Titan X GPU. More detailed qualitative and quantitative results, including success and failure cases, are included in the supplementary material.

### 4.5. Results on SBD Dataset

We also evaluate our model on the SBD dataset, which consists of 5623 training and 5732 validation images, as shown in Tab. 3. Following other works, we only report $AP^r$ results at IoU thresholds of 0.5 and 0.7. However, we provide more detailed results in our supplementary material. Once again, we show significant improvements over other work at high $AP^r$ thresholds. Here, our $AP^r$ at 0.7 improves by 1.5% over the previous state-of-the-art [30]. Note that [30, 37, 20] perform additional post-processing where their results are rescored using an additional object detector. In contrast, our results are obtained by a single forward pass through our network. We have also improved substantially on the $AP^r_{vol}$ measure (3.4%) compared to other works which have reported it. We also used the publicly available source code[1], model and default parameters of MNC [12] to evaluate the "Matching IoU". Our method improves this by 8.3%. This metric is a stricter measure of segmentation performance, and our method, which is based on an initial semantic segmentation and includes a CRF as part of training therefore performs better.

### 4.6. Improvement in Semantic Segmentation

Finetuning our network for instance segmentation, with the loss described in Sec. 3.4 improves semantic segmentation performance on both the VOC and SBD dataset, as shown in Tab. 4. The improvement is 0.9% on VOC, and 1% on SBD. The tasks of instance segmentation and semantic segmentation are highly related – in fact, instance segmentation can be thought of as a more specific case of semantic segmentation. As a result, finetuning for one task improves the other.

---
[1] https://github.com/daijifeng001/MNC

Table 4: Semantic Segmentation performance before and after finetuning for Instance Segmentation on the VOC and SBD Validation Sets

| Dataset | Mean IoU [%] before Instance finetuning | Mean IoU [%] after Instance finetuning |
| --- | --- | --- |
| VOC | 74.2 | 75.1 |
| SBD | 71.5 | 72.5 |

### 4.7. Results on Cityscapes

Finally, we evaluate our algorithm on the Cityscapes road-scene understanding dataset [9]. This dataset consists of 2975 training images, and the held-out test set consisting of 1525 images are evaluated on an online server. None of the 500 validation images were used for training. We use an initial semantic segmentation subnetwork that is based on the ResNet-101 architecture [59], and all of the instance unary potentials described in Sec. 3.2.

As shown in Tab. 5, our method sets a new state-of-the-art on Cityscapes, surpassing concurrent work [21] and the best previous published work [49] by significant margins.

Table 5: Results on Cityscapes Test Set. Evaluation metrics and results of competing methods obtained from the online server. The "AP" metric of Cityscapes is similar to our $AP^r_{vol}$ metric.

| Method | AP | AP at 0.5 | AP 100m | AP 50m |
| --- | --- | --- | --- | --- |
| **Ours** | **20.0** | **38.8** | **32.6** | **37.6** |
| SAIS [21] | 17.4 | 36.7 | 29.3 | 34.0 |
| DWT [4] | 15.6 | 30.0 | 26.2 | 31.8 |
| InstanceCut [24] | 13.0 | 27.9 | 22.1 | 26.1 |
| Graph Decomp. [29] | 9.8 | 23.2 | 16.8 | 20.3 |
| RecAttend [43] | 9.5 | 18.9 | 16.8 | 20.9 |
| Pixel Encoding [49] | 8.9 | 21.1 | 15.3 | 16.7 |
| R-CNN [9] | 4.6 | 12.9 | 7.7 | 10.3 |

## 5. Conclusion and Future Work

We have presented an end-to-end instance segmentation approach that produces intermediate semantic segmentations, and shown that finetuning for instance segmentation improves our network's semantic segmentations. Our approach differs from other methods which derive their architectures from object detection networks [12, 37, 20] in that our approach is more similar to a semantic segmentation network. As a result, our system produces more accurate and detailed segmentations as shown by our substantial improvements at high $AP^r$ thresholds. Moreover, our system produces segmentation maps naturally, and in contrast to other published work, does not require any post-processing. Finally, our network produces a variable number of outputs, depending on the number of instances in the image. Our future work is to incorporate an object detector into the end-to-end training of our system to create a network that performs semantic segmentation, object detection and instance segmentation jointly. Possible techniques for doing this are suggested by [25] and [39].

**Acknowledgements** We thank Bernardino Romera-Paredes and Stuart Golodetz for insightful discussions and feedback. This work was supported by the EPSRC, Clarendon Fund, ERC grant ERC-2012-AdG 321162-HELIOS, EPRSRC grant Seebibyte EP/M013774/1 and EPSRC/MURI grant EP/N019474/1.

# Appendix

In this supplementary material, we include more detailed qualitative and quantitative results on the VOC and SBD datasets. Furthermore, we also show the runtime of our algorithm.

Figures 7 and 8 show success and failure cases of our algorithm. Figure 9 compares the results of our algorithm to the publicly available model for MNC [12]. Figure 10 compares our results to those of FCIS [31], concurrent work which won the COCO 2016 challenge. Figure 11 presents some qualitative results on the Cityscapes dataset.

Section A shows more detailed results on the VOC dataset. Figure 12 shows a visualisation of our results at different $AP^r$ thresholds, and Tables 7 to 9 show per-class $AP^r$ results at thresholds of 0.5, 0.7 and 0.9.

Section B shows more detailed results on the SBD dataset. Table 6 shows our mean $AP^r$ results at thresholds from 0.5 to 0.9, whilst Tables 10 and 11 show per-class $AP^r$ results at thresholds of 0.7 and 0.5 respectively.

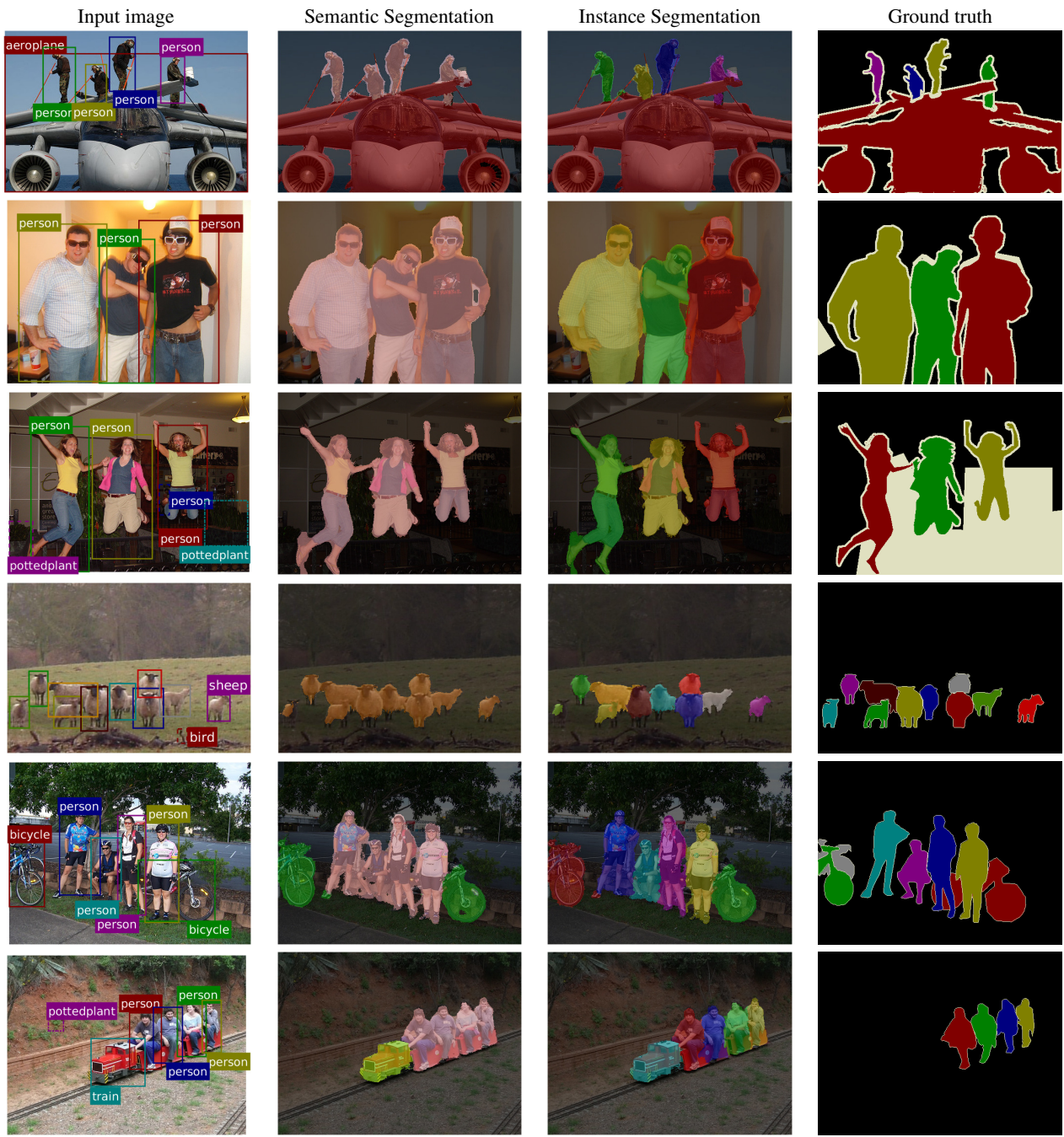

Figure 7: **Success cases of our method.** *First and second row:* Our algorithm can leverage good initial semantic segmentations, and detections, to produce an instance segmentation. *Third row:* Notice that we have ignored three false-positive detections. Additionally, the red bounding box does not completely encompass the person, but our algorithm is still able to associate pixels "outside-the-box" with the correct detection (also applies to row 2). *Fourth row:* Our system is able to deal with the heavily occluded sheep, and ignore the false-positive detection. *Fifth row:* We have not been able to identify one bicycle on the left since it was not detected, but otherwise have performed well. *Sixth row:* Although subjective, the train has not been annotated in the dataset, but both our initial semantic segmentation and object detection networks have identified it. Note that the first three images are from the VOC dataset, and the last three from SBD. Annotations in the VOC dataset are more detailed, and also make more use of the grey "ignore" label to indicate uncertain areas in the image. The first column shows the input image, and the results of our object detector which are another input to our network. Best viewed in colour.

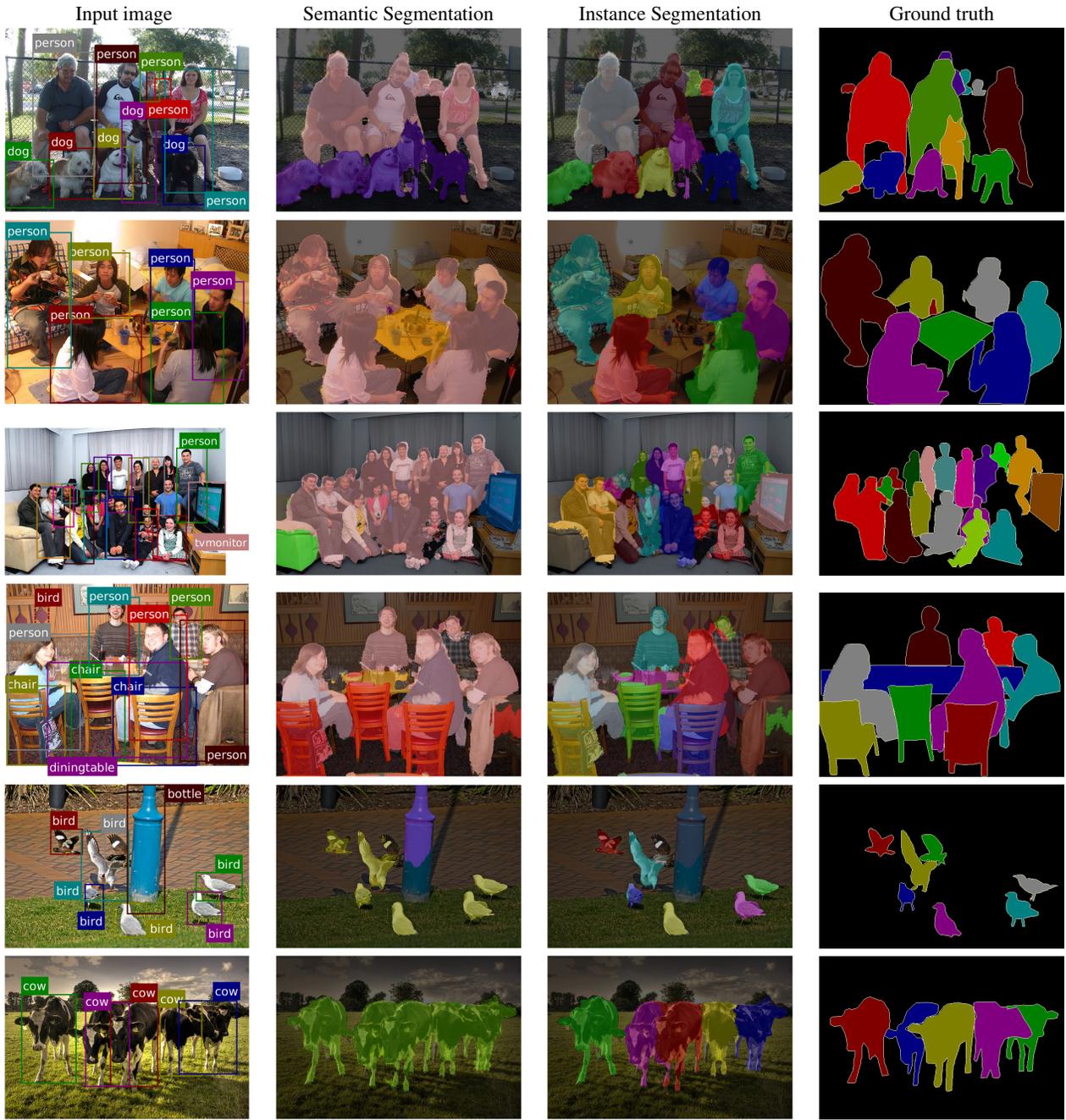

Figure 8: **Failure cases of our method.** *First row:* Both our initial detector, and semantic segmentation system did not identify a car in the background. Additionally, the "brown" person prediction actually consists of two people that have been merged together. This is because the detector did not find the background person. *Second row:* Our initial semantic segmentation identified the table, but it is not there in the Instance Segmentation. This is because there was no "table detection" to associate these pixels with. Using heuristics, we could propose additional detections in cases like these. However, we have not done this in our work. *Third row:* A difficult case where we have segmented most of the people. However, sometimes two people instances are joined together as one person instance. This problem is because we do not have a detection for each person in the image. *Fourth row:* Due to our initial semantic segmentation, we have not been able to segment the green person and table correctly. *Fifth row:* We have failed to segment a bird although it was detected. *Sixth row:* The occluding cows, which all appear similar, pose a challenge, even with our shape priors. The first column shows the input image, and the results of our object detector which are another input to our network. Best viewed in colour.

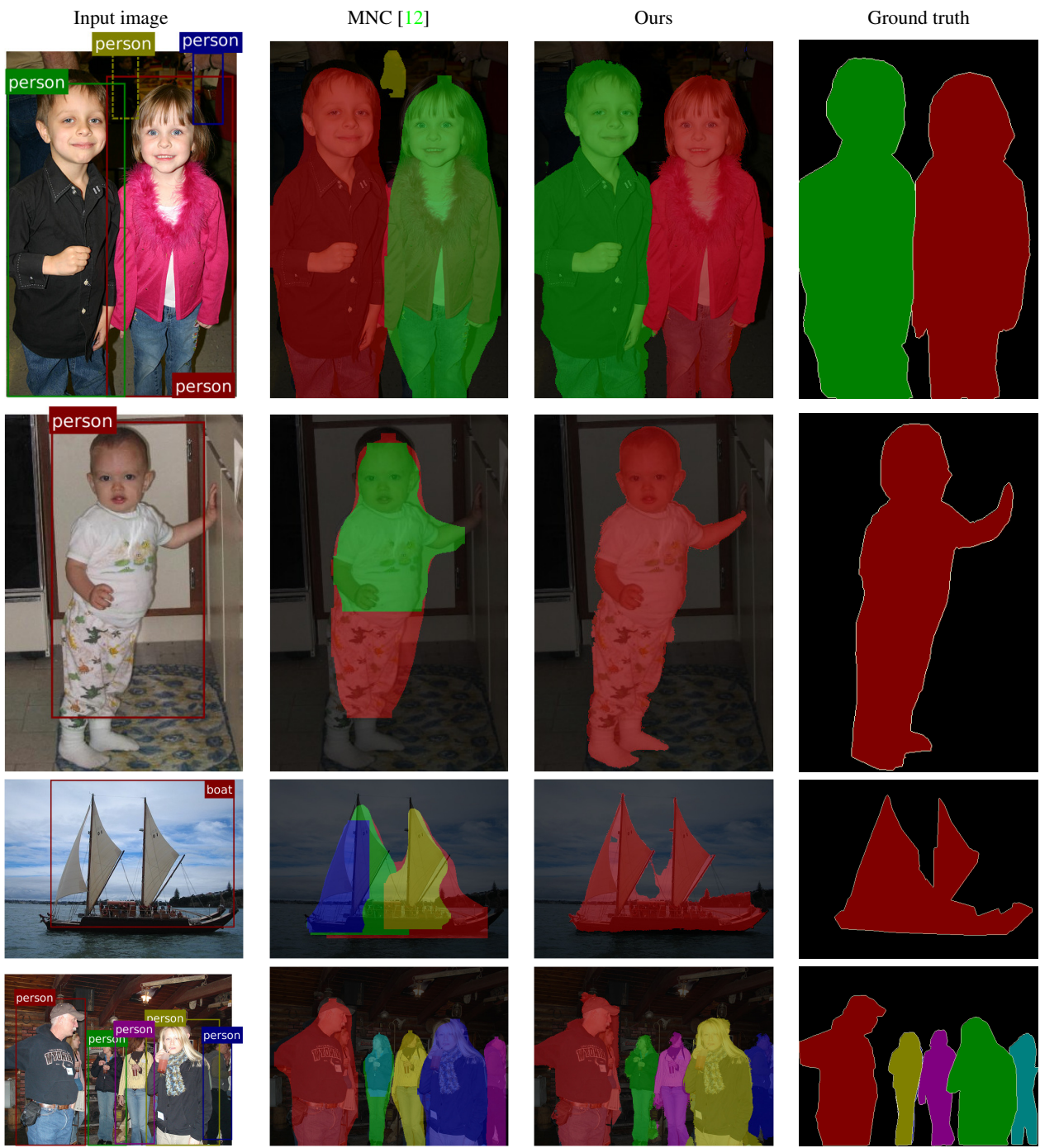

Figure 9: **Comparison to MNC [12]** The above examples emphasise the advantages in our method over MNC [12]. Unlike proposal-based approaches such as MNC, our method can handle false-positive detections, poor bounding box localisation, reasons globally about the image and also produces more precise segmentations due to the initial semantic segmentation module which includes a differentiable CRF. *Row 1* shows a case where MNC, which scores segment-based proposals, is fooled by a false-positive detection and segments an imaginary human (yellow segment). Our method is robust to false-positive detections due to the initial semantic segmentation module which does not have the same failure modes as the detector. *Rows 2, 3 and 4* show how MNC [12] cannot deal with poorly localised bounding boxes. The horizontal boundaries of the red person in Row 2, and light-blue person in Row 4 correspond to the limits of the proposal processed by MNC. Our method, in contrast, can segment "outside the detection bounding box" due to the global instance unary potential (Eq. 4). As MNC does not reason globally about the image, it cannot handle cases of overlapping bounding boxes well, and produces more instances than there actually are. The first column shows the input image, and the results of our object detector which are another input to our network. MNC does not use these detections, but does internally produce box-based proposals which are not shown. Best viewed in colour.

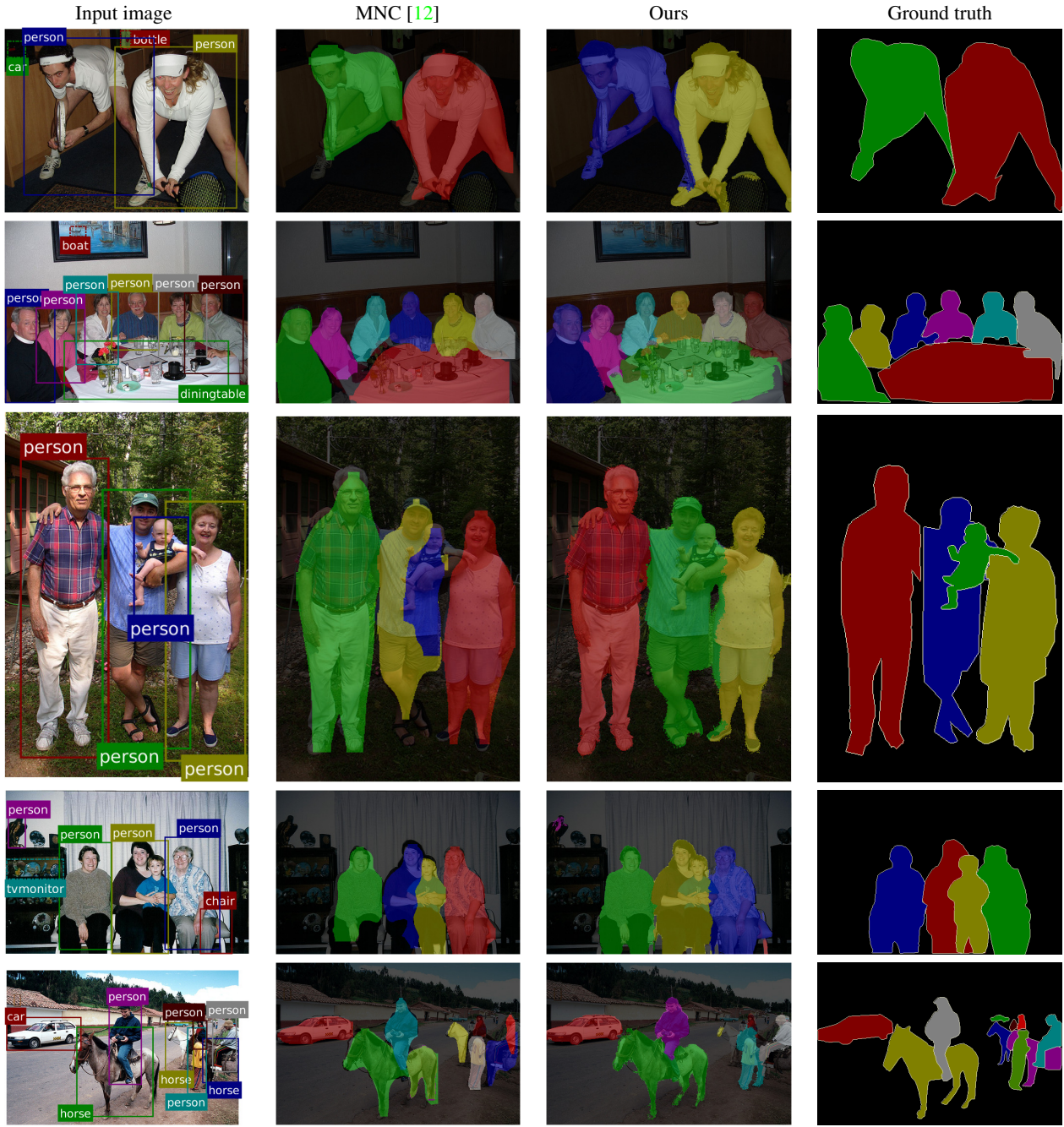

Figure 9 continued: **Comparison to MNC [12]** The above examples show that our method produces more precise segmentations than MNC, that adhere to the boundaries of the objects. However, in Rows 3, 4 and 5, we see that MNC is able to segment instances that our method misses out. In *Row 3*, our algorithm does not segment the baby, although there is a detection for it. This suggests that our shape prior which was formulated to overcome such occlusions could be better. As MNC processes individual instances, it does not have a problem with dealing with small, occluding instances. In *Row 4*, MNC has again identified a person that our algorithm could not. However, this is because we did not have a detection for this person. In *Row 5*, MNC has segmented the horses on the right better than our method. The first column shows the input image, and the results of our object detector which are another input to our network. MNC does not use these detections, but does internally produce box-based proposals which are not shown. We used the publicly available code, models and default parameters of MNC to produce this figure. Best viewed in colour.

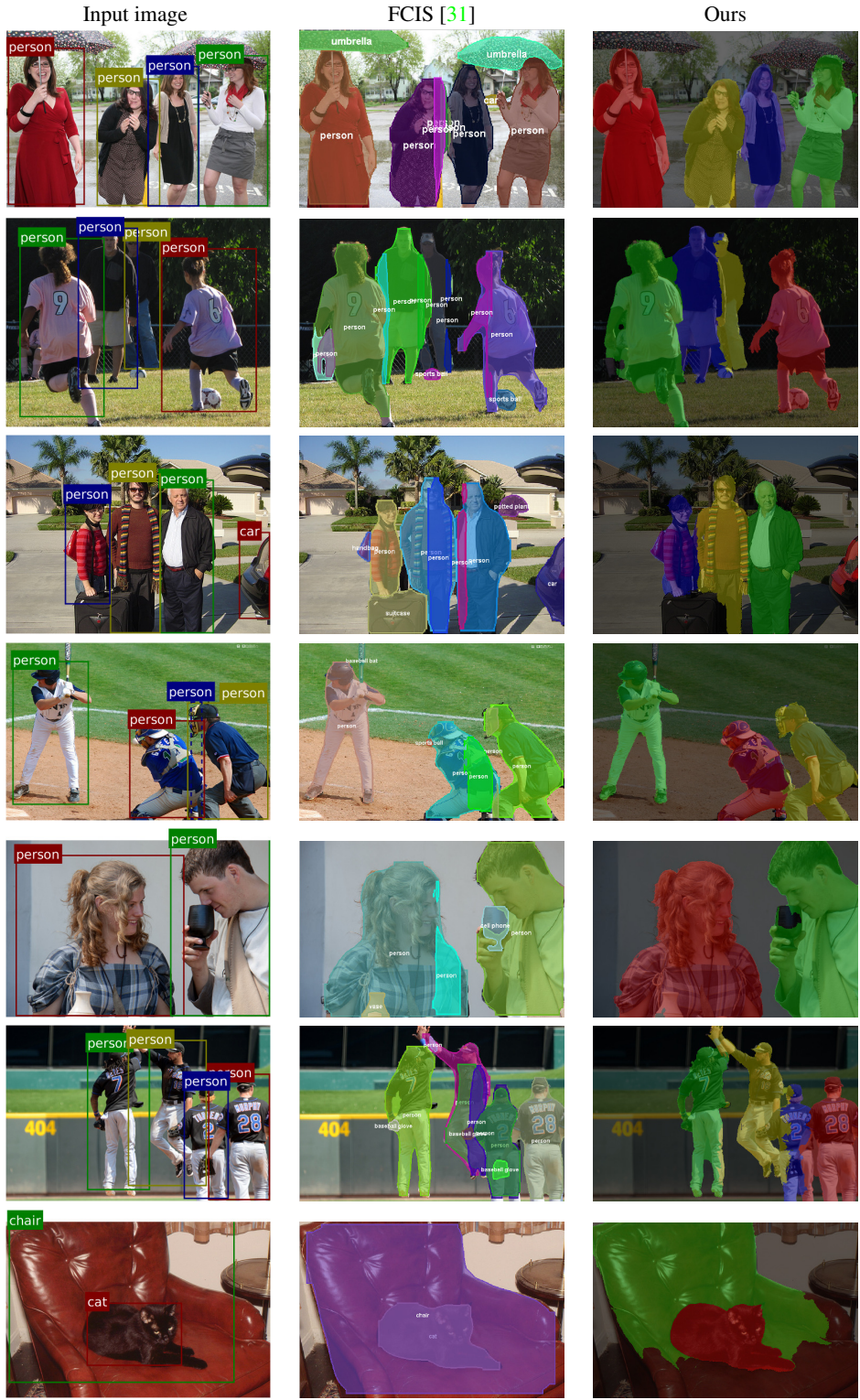

Figure 10: **Comparison to FCIS [31]** The above images compare our method to the concurrent work, FCIS [31], which was trained on COCO [35] and won the COCO 2016 challenge. Unlike proposal-based methods such as FCIS, our method can handle false-positive detections and poor bounding-box localisation. Furthermore, as our method reasons globally about the image, one pixel can only be assigned to a single instance, which is not the case with FCIS. Our method also produces more precise segmentations, as it includes a differentiable CRF, and it is based off a semantic segmentation network. The results of FCIS are obtained from their publicly available results on the COCO test set (https://github.com/daijifeng001/TA-FCN). Note that FCIS is trained on COCO, and our model is trained on Pascal VOC which does not have as many classes as COCO, such as "umbrella" and "suitcase" among others. As a result, we are not able to detect these objects. The first column shows the input image, and the results of our object detector which are another input to our network. FCIS does not use these detections, but does internally produce proposals which are not shown. Best viewed in colour.

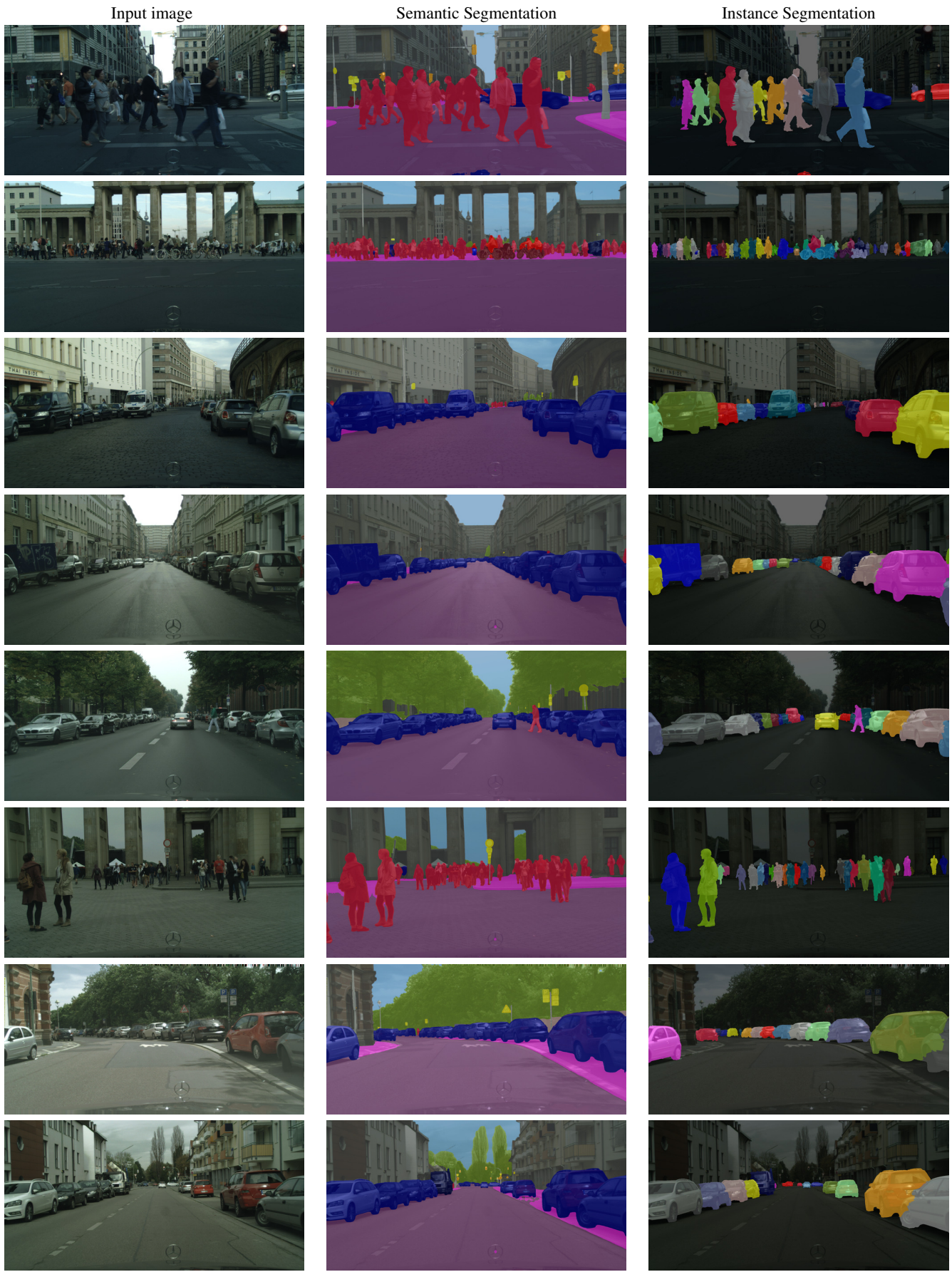

Figure 11: **Sample results on the Cityscapes dataset** The above images show how our method can handle the large numbers of instances present in the Cityscapes dataset. Unlike other recent approaches, our algorithm can deal with objects that are not continuous – such as the car in the first row which is occluded by a pole. Best viewed in colour.

## A. Detailed results on the VOC dataset

Figure 12 shows a visualisation of the $AP^r$ obtained by our method for each class across nine different thresholds. Each "column" of Fig. 12 corresponds to the $AP^r$ for each class at a given IoU threshold. It is therefore an alternate representation for the results tables (Tables 7 to 9). We can see that our method struggles with classes such as "bicycle", "chair", "dining table" and "potted plant". This may be explained by the fact that current semantic segmentation systems (including ours) struggle with these classes. All recent methods on the Pascal VOC leaderboard [2] obtain an IoU for these classes which is lower than the mean IoU for all classes. In fact the semantic segmentation IoU for the "chair" class is less than half of the mean IoU for all the classes for 16 out of the 20 most recent submissions on the VOC leaderboard at the time of writing.

Tables 7 to 9 show per-class instance segmentation results on the VOC dataset, at IoU thresholds of 0.9, 0.7 and 0.5 respectively. At an IoU threshold of 0.9, our method achieves the highest $AP^r$ for 16 of the 20 object classes. At the threshold of 0.7, we achieve the highest $AP^r$ in 15 classes. Finally, at an IoU threshold of 0.5, our method, MPA 3-scale [37] and PFN [33] each achieve the highest $AP^r$ for 6 categories.

## B. Detailed results on the SBD dataset

Once again, we show a visualisation of the $AP^r$ obtained by our method for each class across nine different thresholds (Fig. 13). The trend is quite similar to the VOC dataset in that our algorithm struggles on the same object classes ("chair", "dining table", "potted plant", "bottle"). Note that our $AP^r$ for the "bicycle" class has improved compared to the VOC dataset. This is probably because the VOC dataset has more detailed annotations. In the VOC dataset, each spoke of a bicycle's wheel is often labelled, whilst in SBD, the entire wheel is labelled as a single circle with the "bicycle" label. Therefore, the SBD dataset's coarser labelling makes it easier for an algorithm to perform well on objects with fine details.

Table 6 shows our mean $AP^r$ over all classes at thresholds ranging from 0.5 to 0.9. Our $AP^r$ at 0.9 is low compared to the result which we obtained on the VOC dataset. This could be for a number of reasons: As the SBD dataset is not as finely annotated as the VOC dataset, it might not be suited for measuring the $AP^r$ at such high thresholds. Additionally, the training data is not as good for training our system which includes a CRF and is therefore able to delineate sharp boundaries. Finally, as the SBD dataset has 5732 validation images (compared to the 1449 in VOC), it leaves less data for pretraining our initial semantic segmentation module. This may hinder our network in being able to produce precise segmentations.

Table 6: Comparison of Instance Segmentation performance at multiple $AP^r$ thesholds on the VOC 2012 Validation Set

| Method | $AP^r$ | | | | | $AP^r_{vol}$ |
|---|---|---|---|---|---|---|
|  | 0.5 | 0.6 | 0.7 | 0.8 | 0.9 |  |
| Ours (piecewise) | 59.1 | 51.9 | 42.1 | 29.4 | 12.0 | 52.3 |
| Ours (end-to-end ) | **62.0** | **54.0** | **44.8** | **32.3** | **13.8** | **55.4** |

Tables 10 and 11 show per-class instance segmentation results on the SBD dataset, at IoU thresholds of 0.7 and 0.5 respectively. We can only compare results at these two thresholds since these are the only thresholds which other work has reported.

---

[2]http://host.robots.ox.ac.uk:8080/leaderboard/displaylb.php?challengeid=11&compid=6

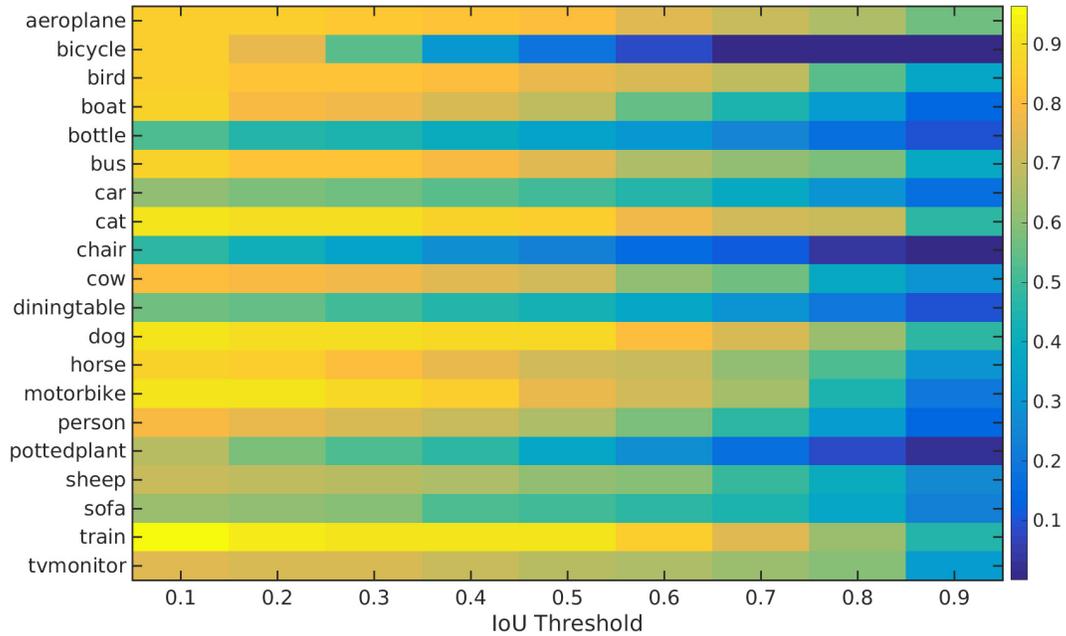

Figure 12: A visualisation of the $AP^r$ obtained for each of the 20 classes on the VOC dataset, at nine different IoU thresholds. The x-axis represents the IoU threshold, and the y-axis each of the Pascal classes. Therefore, each "column" of this figure corresponds to the $AP^r$ per class at a particular threshold, and is thus an alternate representation to the results tables. Best viewed in colour.

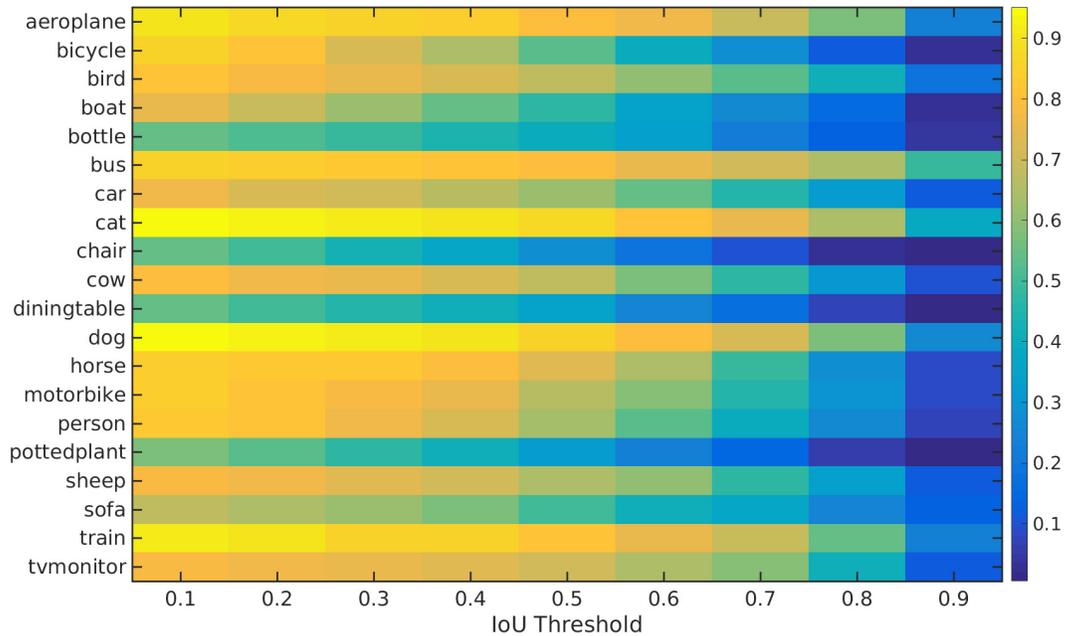

Figure 13: A visualisation of the $AP^r$ obtained for each of the 20 classes on the SBD dataset, at nine different IoU thresholds. The x-axis represents the IoU threshold, and the y-axis each of the Pascal classes. Therefore, each "column" of this figure corresponds to the $AP^r$ per class at a particular threshold, and is thus an alternate representation to the results tables. Best viewed in colour.

Table 7: Comparison of mean $AP^r$, achieved by different published methods, at an IoU threshold of 0.9, for all twenty classes in the VOC dataset.

| Method | Mean $AP^r$(%) | aero-plane | bike | bird | boat | bot-tle | bus | car | cat | chair | cow | table | dog | horse | mbike | per-son | plant | sheep | sofa | train | tv |
|---|---|---|---|---|---|---|---|---|---|---|---|---|---|---|---|---|---|---|---|---|---|
| **Our method** | **25.1** | **56.6** | 0.03 | **36.8** | **14.4** | 9.9 | 39.0 | **17.2** | **47.1** | **1.3** | **29.0** | 9.5 | **47.2** | **29.8** | **20.0** | **14.8** | **2.3** | **25.9** | **23.8** | **45.7** | **32.3** |
| MPA 3-scale [37] | 18.5 | – | – | – | – | – | – | – | – | – | – | – | – | – | – | – | – | – | – | – | – |
| MPA 1-scale [37] | 17.3 | – | – | – | – | – | – | – | – | – | – | – | – | – | – | – | – | – | – | – | – |
| Arnab et al. [3] | 20.1 | 43.7 | 0.03 | 30.0 | 13.2 | **11.4** | **47.3** | 10.9 | 34.5 | 0.7 | 19.6 | **12.1** | 35.6 | 24.3 | 13.3 | 10.7 | 0.4 | 20.7 | 20.9 | 35.0 | 17.4 |
| PFN [33] | 15.7 | 43.9 | **0.1** | 24.5 | 7.8 | 4.1 | 32.5 | 6.3 | 42.0 | 0.6 | 25.7 | 3.2 | 31.8 | 13.4 | 8.1 | 5.9 | 1.6 | 14.8 | 14.3 | 25.0 | 8.5 |
| Chen et al. [8] | 2.6 | 0.6 | 0 | 0.6 | 0.5 | 4.9 | 9.8 | 1.1 | 8.3 | 0.1 | 1.1 | 1.2 | 1.7 | 0.3 | 0.8 | 0.6 | 0.3 | 0.8 | 7.6 | 4.3 | 6.2 |
| SDS [19] | 0.9 | 0 | 0 | 0.2 | 0.3 | 2.0 | 3.8 | 0.2 | 0.9 | 0.1 | 0.2 | 1.5 | 0 | 0 | 0 | 0.1 | 0.1 | 0 | 2.3 | 0.2 | 5.8 |

Table 8: Comparison of mean $AP^r$, achieved by different published methods, at an IoU threshold of 0.7, for all twenty classes in the VOC dataset.

| Method | Mean $AP^r$(%) | aero-plane | bike | bird | boat | bot-tle | bus | car | cat | chair | cow | table | dog | horse | mbike | per-son | plant | sheep | sofa | train | tv |
|---|---|---|---|---|---|---|---|---|---|---|---|---|---|---|---|---|---|---|---|---|---|
| **Our method** | **48.6** | **69.6** | 1.4 | **68.2** | **45.1** | 25.2 | 61.1 | **38.7** | 72.1 | **11.2** | **56.3** | 30.0 | **73.3** | 60.7 | **64.3** | **46.8** | **17.1** | **49.1** | **44.6** | **75.0** | **62.0** |
| MPA 3-scale [37] | 47.4 | – | – | – | – | – | – | – | – | – | – | – | – | – | – | – | – | – | – | – | – |
| MPA 1-scale [37] | 45.9 | – | – | – | – | – | – | – | – | – | – | – | – | – | – | – | – | – | – | – | – |
| Arnab et al. [3] | 45.4 | 68.9 | 0.84 | 65.1 | 38.3 | **26.3** | **64.7** | 31.8 | 72.7 | 6.7 | 45.4 | **32.9** | 67.9 | 60.0 | 63.7 | 41.1 | 13.4 | 43.9 | 41.1 | 74.6 | 48.1 |
| PFN [33] | 42.5 | 68.5 | **5.6** | 60.4 | 34.8 | 14.9 | 61.4 | 19.2 | **78.6** | 4.2 | 51.1 | 28.2 | 69.6 | **60.7** | 60.5 | 26.5 | 9.8 | 35.1 | 43.9 | 71.2 | 45.6 |
| Chen et al. [8] | 27.0 | 40.8 | 0.07 | 40.1 | 16.2 | 19.6 | 56.2 | 26.5 | 46.1 | 2.6 | 25.2 | 16.4 | 36.0 | 22.1 | 20.0 | 22.6 | 7.7 | 27.5 | 19.5 | 47.7 | 46.7 |
| SDS [19] | 21.3 | 17.8 | 0 | 32.5 | 7.2 | 19.2 | 47.7 | 22.8 | 42.3 | 1.7 | 18.9 | 16.9 | 20.6 | 14.4 | 12.0 | 15.7 | 5.0 | 23.7 | 15.2 | 40.5 | 51.4 |

Table 9: Comparison of mean $AP^r$, achieved by different published methods, at an IoU threshold of 0.5, for all twenty classes in the VOC dataset.

| Method | Mean $AP^r$(%) | aero-plane | bike | bird | boat | bot-tle | bus | car | cat | chair | cow | table | dog | horse | mbike | per-son | plant | sheep | sofa | train | tv |
|---|---|---|---|---|---|---|---|---|---|---|---|---|---|---|---|---|---|---|---|---|---|
| **Our method** | 61.7 | 80.2 | **19.3** | **76.4** | **69.0** | 35.3 | 74.5 | 50.8 | 84.5 | 22.8 | 70.9 | 43.3 | **87.7** | 71.3 | 76.2 | 65.6 | 37.2 | **61.3** | 50.3 | **90.5** | 67.2 |
| MPA 3-scale [37] | **62.1** | 79.7 | 11.5 | 71.6 | 54.6 | **44.7** | **80.9** | **62.0** | 85.4 | **26.5** | 64.5 | 46.6 | 87.6 | 71.7 | **77.9** | **72.1** | 48.8 | 57.4 | 48.8 | 78.9 | 70.8 |
| MPA 1-scale [37] | 60.3 | 79.2 | 13.4 | 71.6 | 59.0 | 41.5 | 73.8 | 52.3 | 87.3 | 23.3 | 61.2 | 42.5 | 83.1 | 70.0 | 77.0 | 67.6 | **50.7** | 56.0 | 45.9 | 80.0 | 70.5 |
| Arnab et al. [3] | 58.4 | **80.4** | 7.9 | 74.4 | 59.8 | 32.7 | 76.6 | 39.6 | 84.6 | 19.3 | 62.7 | 44.1 | 81.0 | 74.7 | 72.0 | 58.6 | 32.0 | 59.6 | 50.5 | 87.4 | 68.4 |
| PFN [33] | 58.7 | 76.4 | 15.6 | 74.2 | 54.1 | 26.3 | 73.8 | 31.4 | **92.1** | 17.4 | **73.7** | **48.1** | 82.2 | **81.7** | 72.0 | 48.4 | 23.7 | 57.7 | **64.4** | 88.9 | **72.3** |
| Chen et al. [8] | 46.3 | 63.6 | 0.3 | 61.5 | 43.9 | 33.8 | 67.3 | 46.9 | 74.4 | 8.6 | 52.3 | 31.3 | 63.5 | 48.8 | 47.9 | 48.3 | 26.3 | 40.1 | 33.5 | 66.7 | 67.8 |
| SDS [19] | 43.8 | 58.8 | 0.5 | 60.1 | 34.4 | 29.5 | 60.6 | 40.0 | 73.6 | 6.5 | 52.4 | 31.7 | 62.0 | 49.1 | 45.6 | 47.9 | 22.6 | 43.5 | 26.9 | 66.2 | 66.1 |

Table 10: Comparison of mean $AP^r$, achieved by different published methods, at an IoU threshold of **0.7**, for all twenty classes in the SBD dataset.

| Method | Mean $AP^r$(%) | aero-plane | bike | bird | boat | bot-tle | bus | car | cat | chair | cow | table | dog | horse | mbike | per-son | plant | sheep | sofa | train | tv |
|---|---|---|---|---|---|---|---|---|---|---|---|---|---|---|---|---|---|---|---|---|---|
| **Our method** | **44.8** | **69.0** | 27.4 | **52.7** | **26.4** | 22.4 | 70.3 | 46.0 | **74.7** | 9.6 | 46.8 | **16.9** | **71.6** | 48.4 | 46.3 | **40.3** | 14.8 | 47.6 | **36.5** | **69.7** | **58.2** |
| IIS sp, rescore [30] | 43.3 | 61.9 | **35.1** | 44.4 | **26.4** | **29.6** | **74.0** | **48.7** | 66.8 | **10.9** | **48.4** | 13.6 | 64.0 | **53.0** | **46.8** | 33.0 | **19.0** | **51.0** | 23.7 | 62.2 | 53.9 |
| IIS raw [30] | 38.7 | 61.8 | 31.5 | 42.0 | 22.0 | 22.7 | 72.4 | 44.8 | 65.4 | 7.2 | 37.6 | 10.4 | 60.4 | 39.6 | 41.9 | 32.5 | 12.0 | 40.9 | 19.9 | 58.8 | 50.8 |

Table 11: Comparison of mean $AP^r$, achieved by different published methods, at an IoU threshold of **0.5**, for all twenty classes in the SBD dataset.

| Method | Mean $AP^r$(%) | aero-plane | bike | bird | boat | bot-tle | bus | car | cat | chair | cow | table | dog | horse | mbike | per-son | plant | sheep | sofa | train | tv |
|---|---|---|---|---|---|---|---|---|---|---|---|---|---|---|---|---|---|---|---|---|---|
| **Our method** | 62.0 | **80.3** | 52.8 | 68.5 | 47.4 | 39.5 | 79.1 | 61.5 | **87.0** | 28.1 | **68.3** | **35.5** | **86.1** | 73.9 | 66.1 | 63.8 | 32.9 | 65.3 | **50.4** | **81.4** | **71.4** |
| IIS sp, rescore [30] | **63.6** | 79.2 | **67.9** | **70.0** | **47.9** | **45.3** | **81.6** | **68.8** | 84.1 | **30.4** | 65.5 | 31.8 | 83.6 | **75.5** | **74.5** | **66.6** | **37.7** | **70.6** | 44.7 | 77.7 | 68.7 |
| IIS raw [30] | 60.1 | 77.3 | 65.3 | 65.5 | 42.5 | 35.4 | 80.3 | 62.2 | 83.9 | 27.2 | 61.6 | 32.4 | 82.3 | 70.9 | 71.4 | 63.1 | 31.3 | 63.6 | 44.9 | 78.3 | 62.4 |